\documentclass{article} 
\usepackage{iclr2020_conference,times}

\usepackage{amsmath,amsfonts,bm}

\def\eqref#1{equation~\ref{#1}}

\def\1{\bm{1}}

\DeclareMathAlphabet{\mathsfit}{\encodingdefault}{\sfdefault}{m}{sl}
\SetMathAlphabet{\mathsfit}{bold}{\encodingdefault}{\sfdefault}{bx}{n}

\usepackage{microtype}
\usepackage{hyperref}
\usepackage{url}
\usepackage{graphicx}
\usepackage{subfigure}
\usepackage{booktabs} 
\usepackage{amsmath}
\usepackage[ruled,vlined]{algorithm2e}
\usepackage{algorithmic}

\title{Accelerating Meta-Learning by Sharing \\Gradients}

\author{Oscar Chang, Hod Lipson \\
\texttt{\{oscar.chang,hod.lipson\}@columbia.edu} \\
}

\iclrfinalcopy 
\begin{document}

\maketitle

\begin{abstract}
The success of gradient-based meta-learning is primarily attributed to its ability to leverage related tasks to learn task-invariant information. However, the absence of interactions between different tasks in the inner loop leads to task-specific over-fitting in the initial phase of meta-training. While this is eventually corrected by the presence of these interactions in the outer loop, it comes at a significant cost of slower meta-learning. To address this limitation, we explicitly encode task relatedness via an inner loop regularization mechanism inspired by multi-task learning. Our algorithm shares gradient information from previously encountered tasks as well as concurrent tasks in the same task batch, and scales their contribution with meta-learned parameters. We show using two popular few-shot classification datasets that \emph{gradient sharing} enables meta-learning under bigger inner loop learning rates and can accelerate the meta-training process by up to 134\%.
\end{abstract}

\section{Introduction}\label{introduction}
Like conventional machine learning, meta-learning algorithms can be prone to the risk of over-fitting. Over-fitting in meta-learning though can occur at both the level of the outer and the inner loop. Much prior work has dealt with the outer loop over-fitting to tasks in the meta-training distribution \citep{jamal2019task,khodak2019provable,rajeswaran2019meta,yin2020metalearning}, but little attention has been paid towards the inner loop over-fitting to task-specific training data points.

During the initial phase of meta-training, the scarce number of data points in each task, especially for few-shot learning setups, inevitably causes the over-fitting of comparatively much bigger neural network models. To counter this, meta-learning methods meta-learn the initialized weights of the inner loop learner as a parameter \citep{finn2017model}. By pooling information across different tasks in the meta-training distribution using the outer loop, the initialization eventually picks up task invariant information and gravitates towards a good basin of attraction that reduces the tendency for the inner loop learner to over-fit \citep{guiroy2019towards}. Hence, unlike conventional or outer loop over-fitting, inner loop over-fitting does not always pose a problem to the generalization ability of the meta-learner.

Nevertheless, limiting the interaction between tasks to take place only through the outer loop iteration significantly slows down the convergence of meta-learning. At the start of meta-training, the over-fitting of the inner loop learners causes them to suffer high test losses. Their parameters correspondingly fail to encode task specific information, reducing the signal available to the meta-test loss and thus, the outer loop. This problem is sustained until the model progresses towards more meaningful solutions in the inner loop, causing a significant number of wasteful initial model updates in the outer loop. In meta-learning, this issue is further exacerbated by the inordinate computational expense of a model update, which scales with the number of data points within each task, the number of tasks, the number of operations used by each inner loop learner, and the ultimate need to backpropagate through all of that.

\begin{figure*}[t]
\vskip 0.05in
\begin{center}
\centerline{\includegraphics[width=\textwidth]{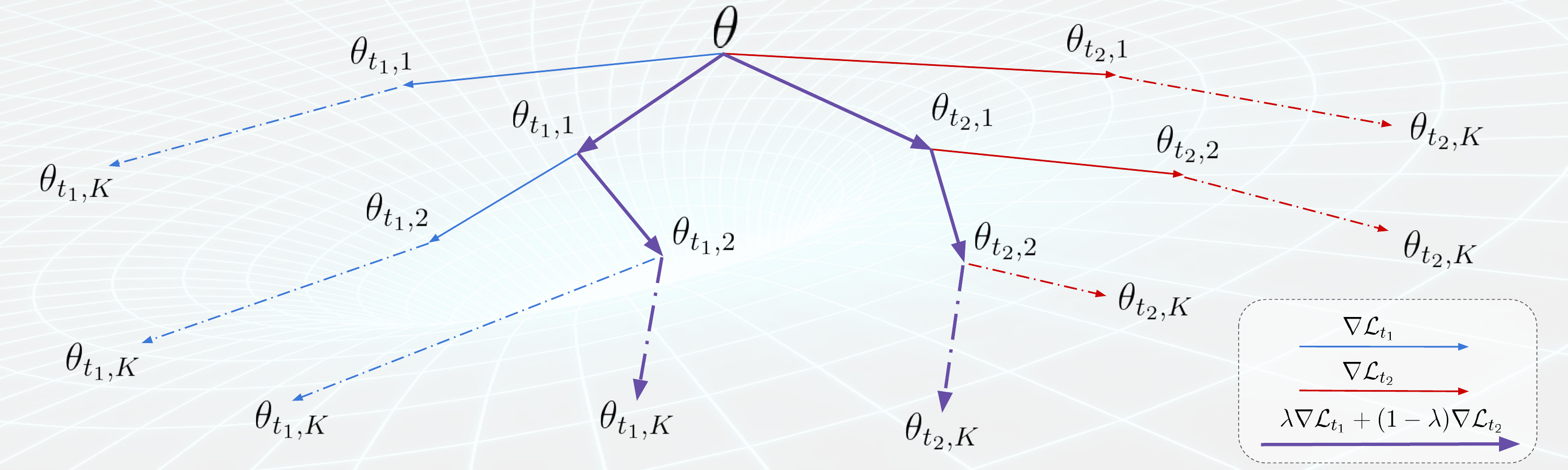}}
\caption{The arrows represent gradient steps taken within the inner loop of meta-learning, with $\theta_{t,k}$ denoting the version of the model after $k$ training steps on task $t$. There are no across-task interactions in the inner loop, causing task-specific over-fitting. This is especially so at the beginning of meta-training, before a good initialization $\theta$ has been meta-learned. The inner loop learning process can be regularized with gradients shared from related tasks.}
\label{fig:grad_share_graphics}
\end{center}
\vskip -0.3in
\end{figure*}

\textbf{Our Contribution} We propose an inner loop regularization mechanism inspired by multi-task learning \citep{caruana1997multitask,baxter1997bayesian} called \emph{gradient sharing}.  Historically, multi-task learning was a predominant approach to leveraging multiple related tasks to learn task-invariant information. Despite this common objective, the rapid development of meta-learning has occurred independently from the vast multi-task learning literature. \textbf{The surprising insight from our work is that the two fields complement each other in a synergistic way.} On one hand, sharing information across tasks in the inner loop via multi-task learning significantly reduces over-fitting. On the other hand, the outer loop can be recruited to meta-learn extra parameters so as to avoid the traditional pitfalls of multi-task learning like imbalanced task combinations. Our proposed method works by sharing gradient information obtained from both previously encountered and concurrent tasks, and scales their contribution with meta-learned parameters. Through extensive experiments on two popular few-shot image classification datasets, we show that gradient sharing accelerates the meta-training process by up to 134\%, and enables meta-learning that is robust to bigger inner loop learning rates while achieving comparable or better meta-test performance. Accelerating meta-training is a key step towards unleashing its full potential, empowering practitioners to use more complex inner loop learners that would have otherwise been intractable.   

\section{Gradient Sharing}
\label{section:gradient_sharing}
Gradient sharing augments the standard MAML inner loop with a meta-learned regularizer that shares gradient information from related tasks and is parametrized by $\boldsymbol{m} \in \mathbb{R}^K, \boldsymbol{\lambda} \in \mathbb{R}^K$. $\sigma$ denotes the sigmoid function. At the $k$-th step of the inner loop, we first compute the normalized average gradient across the task batch (Equation \ref{eq:grad_mean}), and use it to update a running mean of task gradients $\hat{g_k}$ with an exponential moving average factor $\sigma(\boldsymbol{m}_k)$ (Equation \ref{eq:running_grad_1}, \ref{eq:running_grad_2}). $\boldsymbol{m}_k$ can be seen as a momentum variable that controls the weight of recent gradient information relative to past gradients. While the model is largely malleable in the early stages of meta-training, it makes sense for $\boldsymbol{m}_k$ to be large so as to keep pace with quickly changing task gradients. By contrast, near the end of meta-training, variations in task gradients can mostly be attributed to sampling noise and thus, a small $\boldsymbol{m}_k$ is needed for stable training. Meta-learning $\boldsymbol{m}_k$ gives the outer loop flexibility to adapt to both scenarios.


Next, the inner loop update is performed with $\Delta_{t,k}$ which is a $\sigma(\boldsymbol{\lambda}_k)$-weighted linear interpolation between the current task gradient $\nabla_{\theta_{t,k-1}} \mathcal{L}^{\text{train}}_{t}(\theta_{t,k-1})$ and the running mean task gradient $\hat{g_k}$ (Equation \ref{eq:inner_update}). $\boldsymbol{\lambda}_k$ is a gating variable that decides the strength of the multi-task learning regularization coming from related task gradients encountered in the current task batch and previously seen tasks. It is also meta-learned, thus allowing both the task distribution and the size of the task batches to determine the appropriate amount of regularization. For simplicity and storage efficiency, we choose to have a single parameter $\boldsymbol{\lambda}_k$ model the relatedness of each task to all other tasks, although it is straightforward to extend gradient sharing to use task-conditioned parameters $\boldsymbol{\lambda}_{t,k}$.

Finally, we combine the inner loop task losses to arrive at the outer loop update (Equation \ref{eq:outer_update}). At meta-test time, the inner loop is regularized using the $\hat{g_k}$ stored during meta-training.


We write the full gradient sharing algorithm in pseudo-code for vanilla MAML during the meta-training and meta-testing phase in Algorithms \ref{alg:grad_share_metatrain} and \ref{alg:grad_share_metatest} respectively. While the pseudo-code is written for vanilla MAML, gradient sharing can be applied in general to any second-order gradient-based meta-learning method (i.e.\ the inner loop has to be differentiable) by using the regularized $\Delta_{t,k}$ in the place of an inner loop task gradient $\nabla_{\theta_{t,k-1}} \mathcal{L}^{\text{train}}_{t}(\theta_{t,k-1})$.

\begin{algorithm}[ht]
  \caption{Gradient Sharing for MAML Meta-Training.}
  \label{alg:grad_share_metatrain}
\begin{algorithmic}
  \STATE Initialize $\theta$, $T$, $K$, $\boldsymbol{m}=\boldsymbol{0}$, $\boldsymbol{\lambda}=\boldsymbol{0}$.
  \FOR{$i=1$ {\bfseries to} $numMetatrainIters$}
  \STATE Sample batch $\mathcal{B}$ with $T$ tasks from meta-training set.
  \STATE Initialize $\theta_{t, 0} = \theta$ for all tasks $t$ in $\mathcal{B}$.
  \STATE \textit{// $K$ is the number of inner loop gradient steps.}
  \FOR{$k=1$ {\bfseries to} $K$}
  \STATE \textit{// Calculate normalized mean of task gradients in $\mathcal{B}$.}
  \vskip -0.1in
  \begin{equation}\label{eq:grad_mean}\boxed{
  g_k = \frac{\sum_{t=1}^T \nabla_{\theta_{t,k-1}} \mathcal{L}^{\text{train}}_{t}(\theta_{t,k-1})}{|| \sum_{t=1}^T \nabla_{\theta_{t,k-1}} \mathcal{L}^{\text{train}}_{t}(\theta_{t,k-1}) ||_2}.
  }\end{equation}
  \STATE \textit{// Calculate running mean gradient statistics $\hat{g_k}$.}
  \IF{$i=1$}
  \STATE
  \vskip -0.15in
  \begin{equation}\label{eq:running_grad_1}\boxed{
      \hat{g_k} = g_k.
  }\end{equation}
  \ELSE
  \STATE
  \vskip -0.15in
  \begin{equation}\label{eq:running_grad_2}\boxed{
      \hat{g_k} = \sigma(\boldsymbol{m}_k) g_k + (1 - \sigma(\boldsymbol{m}_k)) \hat{g_k}.
  }\end{equation}
  \ENDIF
  \FOR{task $t$ {\bfseries in} batch $\mathcal{B}$}
  \STATE \textit{// Apply inner loop update.}
  \vskip -0.1in
  \begin{equation}\label{eq:inner_update}\boxed{
  \begin{split}
  \Delta_{t,k} =&\ \sigma(\boldsymbol{\lambda}_k) \hat{g_k} \ \ + \\
  &\ (1 - \sigma(\boldsymbol{\lambda}_k)) \nabla_{\theta_{t,k-1}} \mathcal{L}^{\text{train}}_{t}(\theta_{t,k-1}).\\
  \theta_{t,k} =&\ \theta_{t,k-1} - \alpha \Delta_{t,k}.
  \end{split}}
  \end{equation}
  \ENDFOR
  \ENDFOR
  \STATE \textit{// Apply outer loop update.}
  \vskip -0.1in
  \begin{equation}\label{eq:outer_update}\boxed{
  \begin{split}
      (\theta',\boldsymbol{m}',\boldsymbol{\lambda}') =&\  (\theta,\boldsymbol{m},\boldsymbol{\lambda}) \ \ - \\
      &\ \frac{\beta}{T} \sum_{t=1}^T \nabla_{(\theta,\boldsymbol{m},\boldsymbol{\lambda})} \mathcal{L}^{\text{test}}_t (\theta_{t,K}).
  \end{split}
  }\end{equation}
  \ENDFOR
\end{algorithmic}
\end{algorithm}

\begin{algorithm}[ht]
  \caption{Gradient Sharing for MAML Meta-Testing.}
  \label{alg:grad_share_metatest}
\begin{algorithmic}
  \FOR{task $t$ {\bfseries in} meta-testing set}
  \STATE Initialize $\theta_{t, 0} = \theta$.
  \FOR{$k=1$ {\bfseries to} $K$}
  \STATE $\Delta_{t,k} = \sigma(\boldsymbol{\lambda}_k) \hat{g_k} \ \ +$ \\
  $\hspace{1.1cm} (1 - \sigma(\boldsymbol{\lambda}_k)) \nabla_{\theta_{t,k-1}} \mathcal{L}^{\text{train}}_{t}(\theta_{t,k-1})$.
  \STATE $\theta_{t,k} = \theta_{t,k-1} - \alpha \Delta_{t,k}$.
  \ENDFOR
  \STATE Evaluate task $t$'s test performance with $\mathcal{L}^{\text{test}}_{t}(\theta_{t,K})$.
  \ENDFOR
\end{algorithmic}
\end{algorithm}

\section{Experimental Results and Discussions}
\label{section:experiments}
In the interest of a comprehensive experimental setup, we study the effects of gradient sharing using two popular few-shot image classification datasets --- the Caltech-UCSD Birds-200-2011 (CUB) dataset \citep{wah2011caltech} and the MiniImagenet dataset \citep{ravi2016optimization} --- in two distinct regimes --- task batches of size $1$ and $5$ --- and across three distinct meta-learning methods --- MAML \citep{finn2017model}, Meta-SGD \citep{li2017meta}, and MAML++ \citep{antoniou2018train}.

\begin{figure*}[ht]
\begin{center}
\centerline{\includegraphics[width=\textwidth]{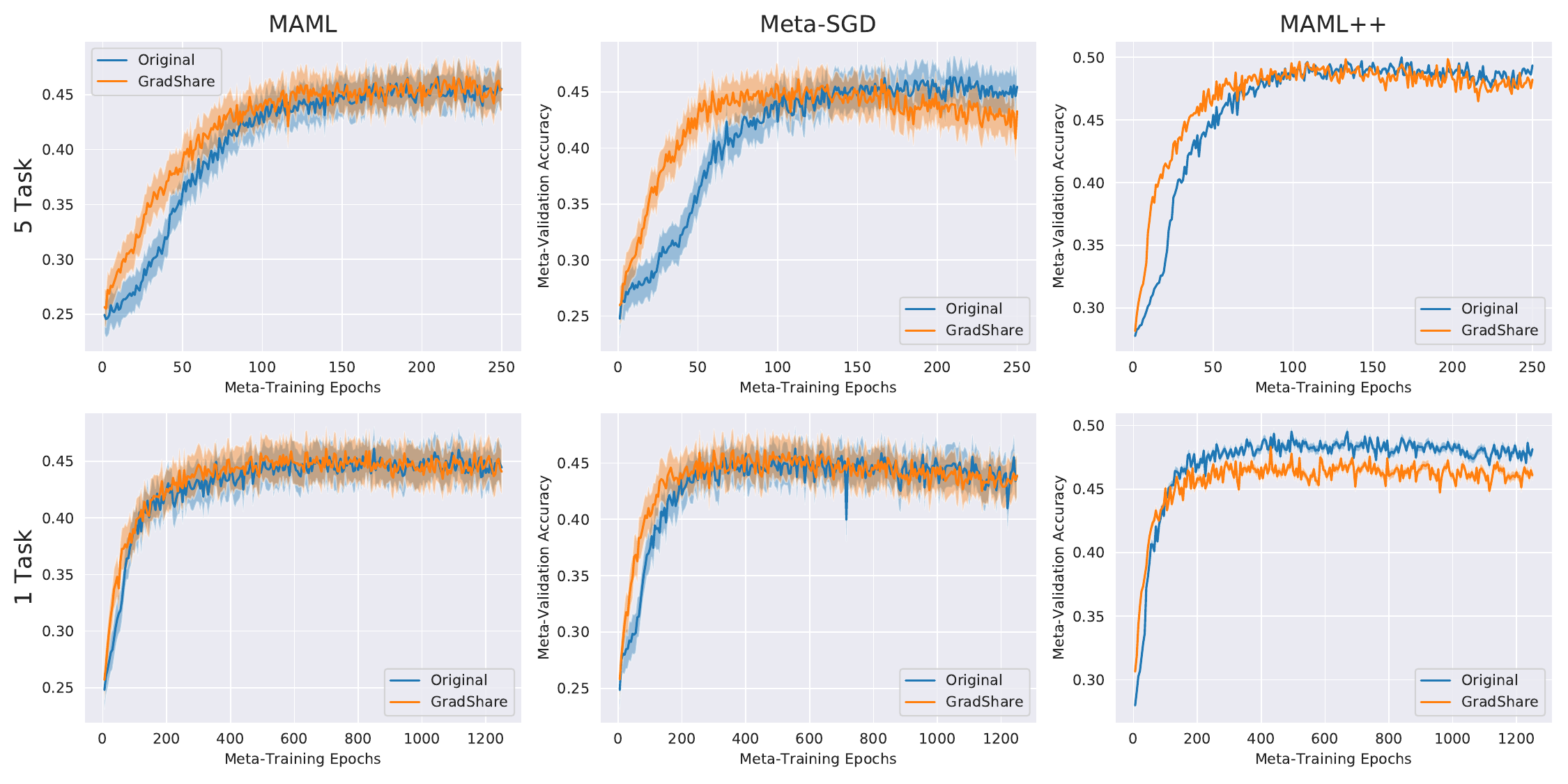}}
\caption{Meta-validation accuracy plots on $5$-way $1$-shot classification on the MiniImagenet dataset. Gradient sharing accelerates meta-learning by reducing inner loop over-fitting in early stage meta-training. The acceleration is stronger when there are other concurrent tasks in the inner loop.}
\label{fig:acc}
\end{center}
\vskip -0.2in
\end{figure*}

We observe in Figure \ref{fig:acc} that in the initial phase of meta-training, gradient sharing results in higher meta-validation performance. This effect is significantly stronger when there are other concurrent tasks in the task batch, due to stronger regularization and smaller variance in task gradients, as we can see by comparing the plots for task batch size $5$ versus $1$. This is clear evidence that gradient sharing is indeed reducing inner loop over-fitting, because it consistently results in higher inner loop test performance early on (Recall that the outer loop loss is a mean of the inner loop test losses).

Achieving superior meta-training performance early on accelerates the overall meta-training process, potentially a speed-up of up to $134\%$ at comparable levels of meta-test accuracy (Table \ref{table:5shot}). To quantify the amount of meta-training acceleration, we use the rate at which the highest meta-validation accuracy is achieved as a proxy. Specifically, we calculate $\text{Speed-up} = \frac{\text{Epoch}_\text{OG}-\text{Epoch}_\text{GS}}{\text{Epoch}_\text{GS}}$ where $\text{Epoch}_\text{OG}$ and $\text{Epoch}_\text{GS}$ is the earliest epoch when the highest meta-validation accuracy is achieved for the original baseline and gradient sharing respectively.

Finally, reducing inner loop over-fitting also enables higher learning rates. Gradient sharing achieves successful meta-training even when the inner loop learners have been initialized with $10$x their learning rate, while the baselines often fail to train at all or experience very sluggish meta-training. Higher inner loop learning rates produces superior meta-test generalization under certain circumstances \citep{li2017meta}, and additional robustness to meta learning hyperparameters is generally very desirable. We direct the interested reader to the Appendix to learn more about the full spectrum of experiments.

\begin{table*}[ht]
\caption{Meta-test accuracy (with $95\%$ confidence intervals) and speed-up for $5$-way $5$-shot classification for the CUB and MiniImagenet datasets. Gradient sharing achieves comparable meta-test accuracy, but often in a fraction of the number of meta-training epochs.}
\label{table:5shot}
\vskip 0.1in
\begin{center}
\begin{small}
\begin{sc}
\begin{tabular}{lcccccc}
\toprule
& \multicolumn{3}{c}{CUB} & \multicolumn{3}{c}{MiniImagenet} \\
\cmidrule(lr){2-4} \cmidrule(lr){5-7}
Method/Tasks & Original & GradShare & Speed-up & Original & GradShare & Speed-up \\
\midrule
MAML/5 & $83.2\pm1.4\%$ & $83.4\pm1.4\%$ & $66\%$ & $67.7\pm1.8\%$ & $67.0\pm1.8\%$ & $1\%$ \\ 
MAML/1 & $82.6\pm1.5\%$ & $82.7\pm1.5\%$ & $44\%$ & $66.4\pm1.8\%$ & $68.3\pm1.8\%$ & $134\%$ \\ 
Meta-SGD/5 & $80.7\pm1.5\%$ & $80.3\pm1.5\%$ & $100\%$ & $67.0\pm1.8\%$ & $67.4\pm1.8\%$ & $61\%$ \\ 
Meta-SGD/1 & $81.6\pm1.5\%$ & $79.6\pm1.6\%$  & $54\%$ & $64.8\pm1.9\%$ & $64.9\pm1.9\%$ & $37\%$ \\ 
MAML++/5 & $72.7\pm1.7\%$ & $73.8\pm1.7\%$ & $42\%$ & 
$68.9\pm1.8\%$ & $69.4\pm1.8\%$ & $26\%$ \\ 
MAML++/1 & $76.1\pm1.7\%$ & $76.5\pm1.6\%$ & $100\%$ & $69.1\pm1.8\%$ & $66.8\pm1.8\%$ & $71\%$ \\
\bottomrule
\end{tabular}
\end{sc}
\end{small}
\end{center}
\end{table*}

\clearpage
\bibliography{iclr2020_conference}
\bibliographystyle{iclr2020_conference}

\clearpage
\appendix
\section*{Appendix}
\section{More Experimental Details}
\subsection{Loading the CUB and MiniImagenet Data}
The CUB dataset consists of $100$ meta-training classes, $50$ meta-validation classes, and $50$ meta-test classes. The MiniImagenet consists of $64$ meta-training classes, $16$ meta-validation classes, and $20$ meta-test classes. Each task involves $5$-way and $1/5$-shot classification on randomly sampled classes using the cross-entropy loss. We use \citet{deleu2019torchmeta}'s Torchmeta dataloader implementation to load the CUB and MiniImagenet datasets for the MAML and Meta-SGD experiments. For the MAML++ experiments, we use \citet{antoniou2018train}'s dataloader \href{https://github.com/AntreasAntoniou/HowToTrainYourMAMLPytorch}{implementation}.

We use $15$ test examples within each task and $600$ evaluation tasks for meta-validation and meta-testing. Following common practice, every image is resized to $84$ by $84$ before being inputted into the model.

\subsection{Model Backbone}
The model used in our experiments is a standard $4$-layer convolutional neural network backbone that is commonly used in the meta-learning literature. We rely on \citet{antoniou2018train}'s \href{ https://github.com/AntreasAntoniou/HowToTrainYourMAMLPytorch/blob/master/meta_neural_network_architectures.py}{implementation}, which has $48$ filters, batch normalization and ReLU activations for each convolutional layer, as well as a max pooling and linear layer before the final softmax. The appropriate flags are set such that the standard MAML backbone is used for the MAML and Meta-SGD experiments, while the version with Per Step Batch Normalization is used for MAML++.

\subsection{Meta-Training and Meta-Testing}
For task batch size $5$, we meta-train on CUB for $150$ epochs and MiniImagenet for $250$ epochs using outer loop Adam \citep{kingma2014adam} with default hyperparameters and inner loop gradient descent with learning rate $0.1$ and $K=5$ steps. For task batch size $1$, we do $5$x as many epochs. Each epoch consists of $1000$ iterations. We did meta-testing using an ensemble of the top $5$ meta-validation accuracy models following the methodology of the MAML++ paper \citep{antoniou2018train}.

\section{More Analysis}
\label{section:more_analysis}
\subsection{Evolution of Momentum $m$ and Lambda $\lambda$ through Meta-Training}
\begin{figure*}[ht!]
\vskip 0.05in
\begin{center}
\centerline{\includegraphics[width=\textwidth]{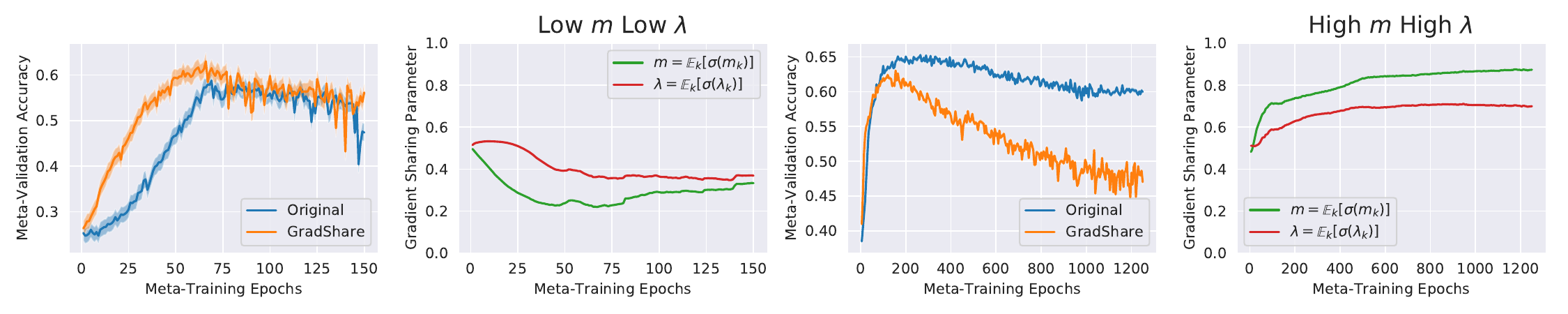}}
\caption{The left two plots show the results of meta-training using gradient sharing on $5$-way $1$-shot classification on CUB using MAML with task batch size $5$. They represent a successful example of gradient sharing with the outer loop meta-learning low values for both $\boldsymbol{m}$ and $\boldsymbol{\lambda}$. The right two plots show meta-training results for $5$-way $5$-shot classification on MiniImagenet using MAML++ with task batch size $1$. They represent a pathological example of gradient sharing with the outer loop meta-learning high values for both $\boldsymbol{m}$ and $\boldsymbol{\lambda}$.}
\label{fig:pathological}
\end{center}
\vskip -0.3in
\end{figure*}

On the left two sub-figures of Figure \ref{fig:pathological}, we observe that accelerated meta-training goes hand in hand with reduced values of the averages of $\boldsymbol{m}_k$ and $\boldsymbol{\lambda}_k$ as meta-training proceeds, which agrees with what we had discussed in Sections \ref{section:gradient_sharing}.

The right two sub-figures of Figure \ref{fig:pathological} show a characteristically different pattern. We observe that the outer loop meta-learns high values of $\boldsymbol{m}_k$ and $\boldsymbol{\lambda}_k$ as meta-training proceeds. High $\boldsymbol{m}_k$ indicates that the store of task gradients $\hat{g_k}$ has not stabilized and recent task gradients are continually overwriting it. High $\boldsymbol{\lambda}_k$ suggests an excessive amount of regularization is being applied; in fact, at the limit of $\sigma(\boldsymbol{\lambda}_k) = 1.0$, the true task gradient $\nabla_{\theta_{t,k-1}} \mathcal{L}^{\text{train}}_{t}(\theta_{t,k-1})$ is completely masked out, effectively making it zero-shot instead of few-shot learning. This pathological phenomenon of high $\boldsymbol{m}_k$ and $\boldsymbol{\lambda}_k$ is congruent with the observed result of gradient sharing exacerbating the original MAML++ baseline's outer loop over-fitting in this case.

The over-fitting of the outer loop has not proven to be a serious issue in our work due to the use of early stopping (since we select the meta-test model using the meta-validation set). However, looking into combinations of outer and inner loop regularization, for example task-conditioned $\boldsymbol{m}_{t,k}$ and $\boldsymbol{\lambda}_{t,k}$, is an important topic for future work.

\section{More Plots}
\label{section:more_plots}
\subsection{Meta-Validation Plots}
\begin{figure*}[ht]
\vskip 0.2in
\begin{center}
\centerline{\includegraphics[width=\textwidth]{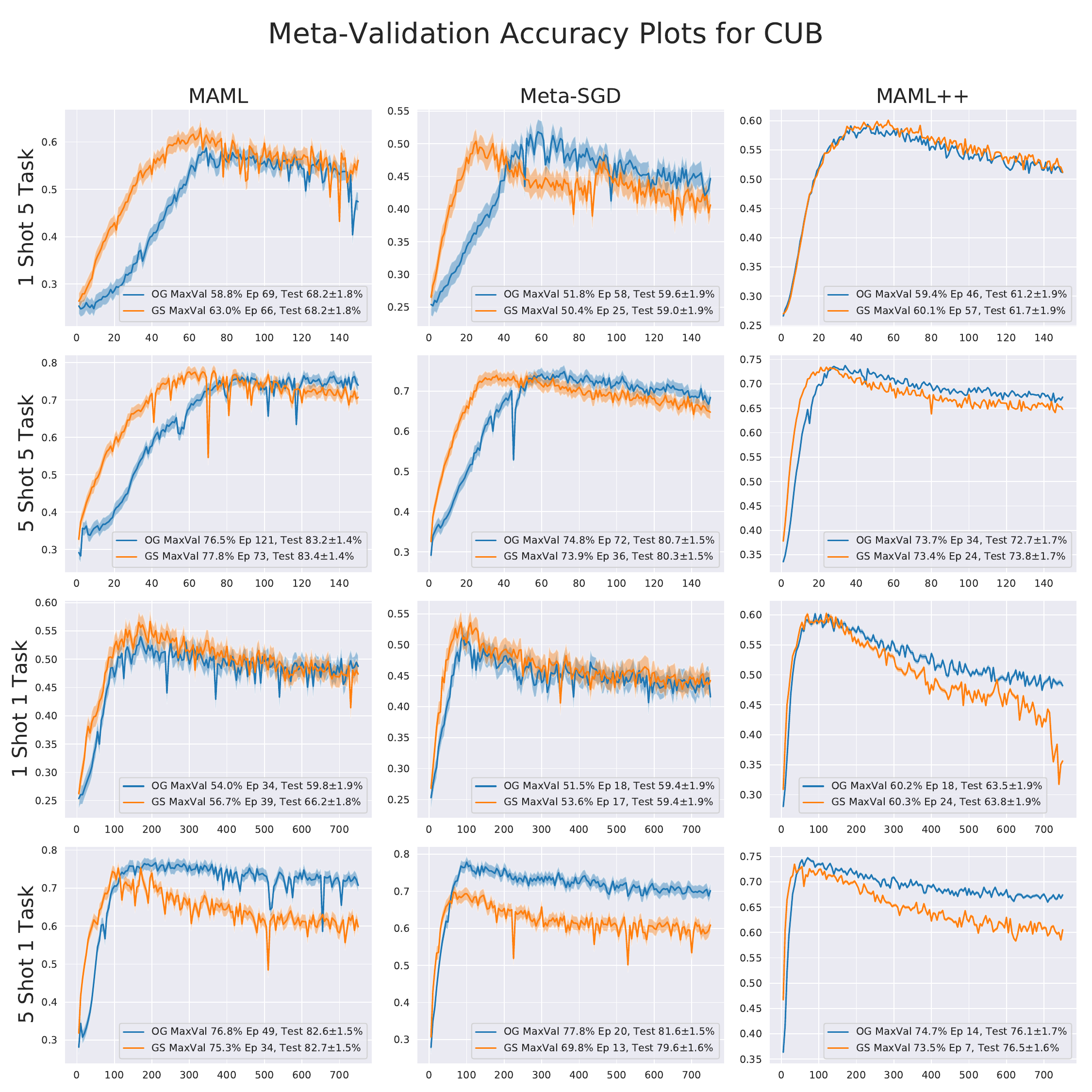}}
\caption{Meta-Validation Accuracy Plots for the CUB dataset. The x axes denote the number of meta-training epochs, the y axes denote the accuracy on the meta-validation set, and the shaded areas denote the $95\%$ standard error confidence interval. In the legend, OG denotes the original baseline meta-learning method, and GS denotes the version with Gradient Sharing. MaxVal [A] Ep [B] denotes that the maximum meta-validation accuracy of [A] was achieved at epoch [B]. Test [C]$\pm$[D] denotes that the meta-test accuracy of [C] was achieved within a $95\%$ confidence interval of [D]. The column headers denote the meta-learning method, while the row headers denote the number of shots and number of tasks in the task batch. All experiments are done in the $5$-way few-shot classification setting, with the meta-test accuracy reported using an ensemble composed of the top $5$ meta-validation accuracy models.}
\label{fig:CUB_val}
\end{center}
\vskip -0.2in
\end{figure*}

\begin{figure*}[ht]
\vskip 0.2in
\begin{center}
\centerline{\includegraphics[width=\textwidth]{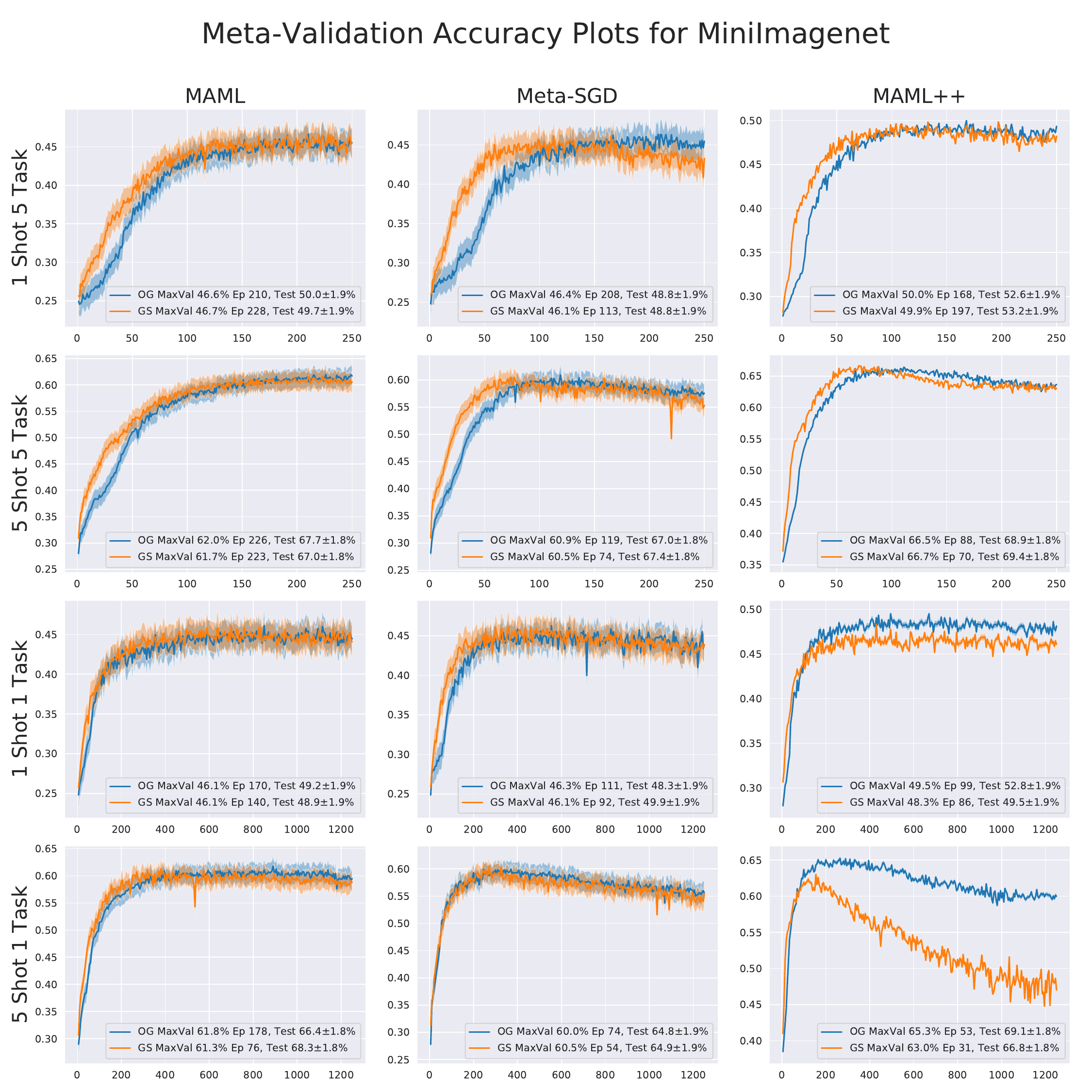}}
\caption{Meta-Validation Accuracy Plots for the MiniImagenet dataset. The x axes denote the number of meta-training epochs, the y axes denote the accuracy on the meta-validation set, and the shaded areas denote the $95\%$ standard error confidence interval. In the legend, OG denotes the original baseline meta-learning method, and GS denotes the version with Gradient Sharing. MaxVal [A] Ep [B] denotes that the maximum meta-validation accuracy of [A] was achieved at epoch [B]. Test [C]$\pm$[D] denotes that the meta-test accuracy of [C] was achieved within a $95\%$ confidence interval of [D]. The column headers denote the meta-learning method, while the row headers denote the number of shots and number of tasks in the task batch. All experiments are done in the $5$-way few-shot classification setting, with the meta-test accuracy reported using an ensemble composed of the top $5$ meta-validation accuracy models.}
\label{fig:MiniImagenet_val}
\end{center}
\vskip -0.2in
\end{figure*}

\begin{figure*}[ht]
\vskip 0.2in
\begin{center}
\centerline{\includegraphics[width=\textwidth]{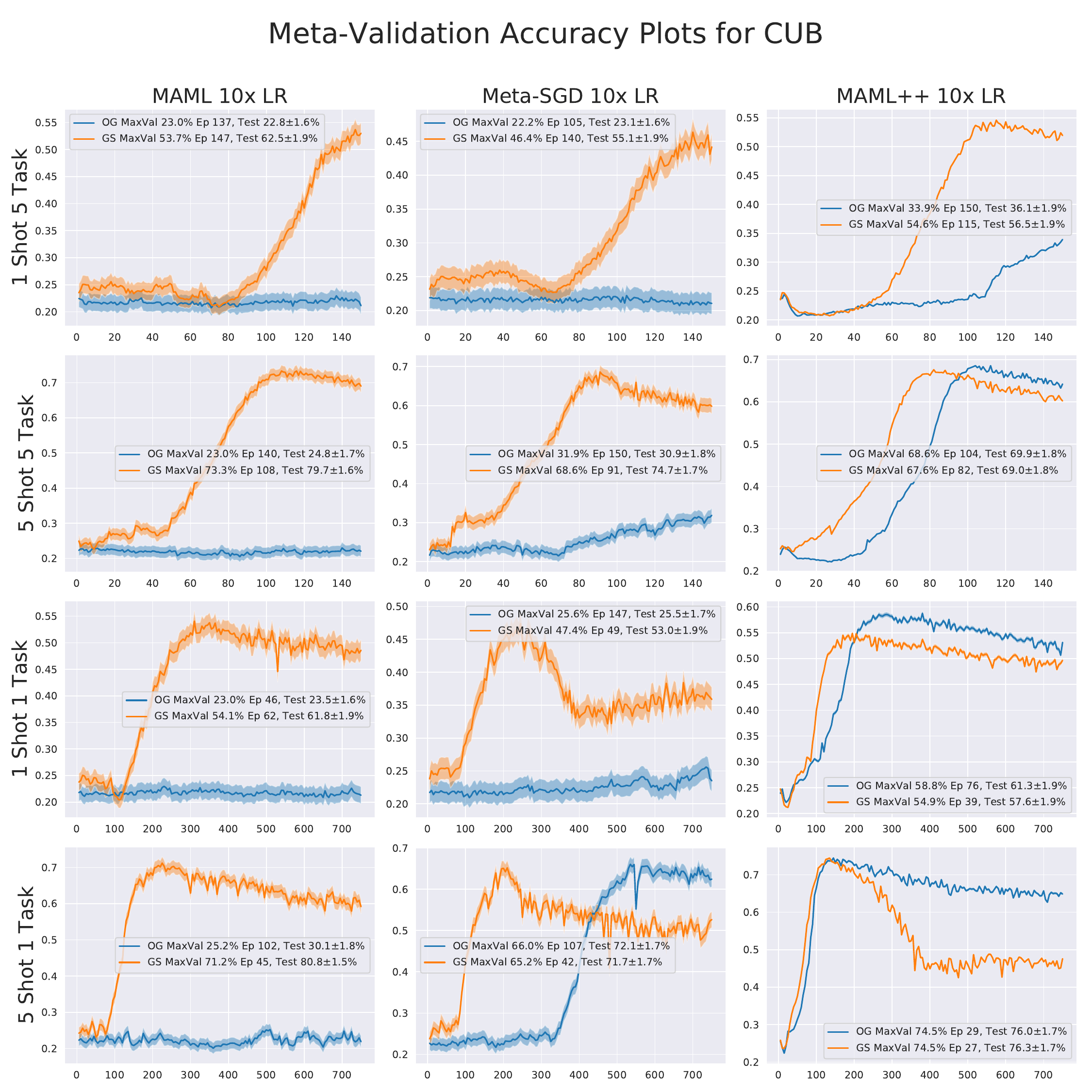}}
\caption{Meta-Validation Accuracy Plots for the CUB dataset with $10$x the Inner Loop Learning Rate. The x axes denote the number of meta-training epochs, the y axes denote the accuracy on the meta-validation set, and the shaded areas denote the $95\%$ standard error confidence interval. In the legend, OG denotes the original baseline meta-learning method, and GS denotes the version with Gradient Sharing. MaxVal [A] Ep [B] denotes that the maximum meta-validation accuracy of [A] was achieved at epoch [B]. Test [C]$\pm$[D] denotes that the meta-test accuracy of [C] was achieved within a $95\%$ confidence interval of [D]. The column headers denote the meta-learning method, while the row headers denote the number of shots and number of tasks in the task batch. All experiments are done in the $5$-way few-shot classification setting, with the meta-test accuracy reported using an ensemble composed of the top $5$ meta-validation accuracy models.}
\label{fig:CUB_biglr_val}
\end{center}
\vskip -0.2in
\end{figure*}

\begin{figure*}[ht]
\vskip 0.2in
\begin{center}
\centerline{\includegraphics[width=\textwidth]{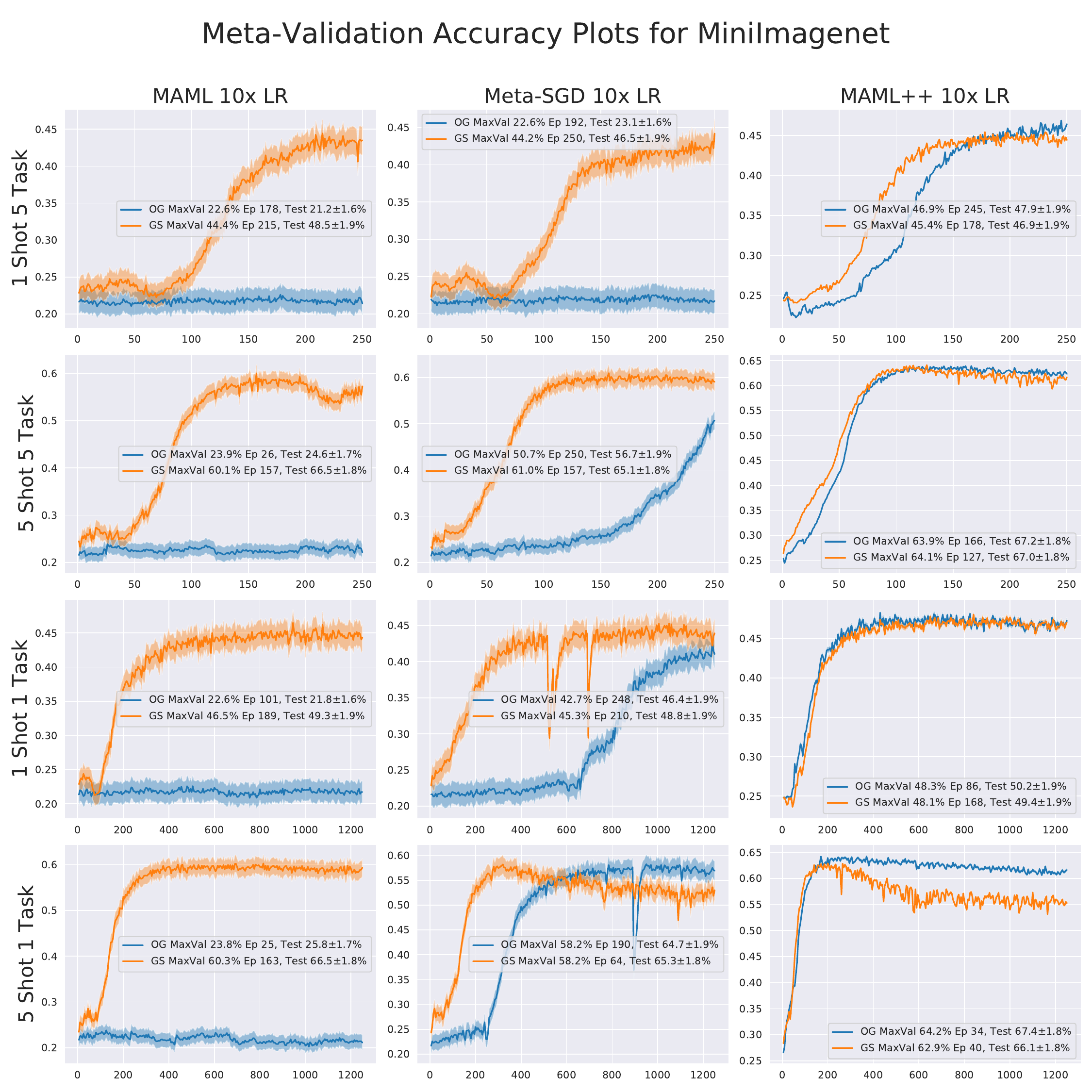}}
\caption{Meta-Validation Accuracy Plots for the MiniImagenet dataset with $10$x the Inner Loop Learning Rate. The x axes denote the number of meta-training epochs, the y axes denote the accuracy on the meta-validation set, and the shaded areas denote the $95\%$ standard error confidence interval. In the legend, OG denotes the original baseline meta-learning method, and GS denotes the version with Gradient Sharing. MaxVal [A] Ep [B] denotes that the maximum meta-validation accuracy of [A] was achieved at epoch [B]. Test [C]$\pm$[D] denotes that the meta-test accuracy of [C] was achieved within a $95\%$ confidence interval of [D]. The column headers denote the meta-learning method, while the row headers denote the number of shots and number of tasks in the task batch. All experiments are done in the $5$-way few-shot classification setting, with the meta-test accuracy reported using an ensemble composed of the top $5$ meta-validation accuracy models.}
\label{fig:MiniImagenet_biglr_val}
\end{center}
\vskip -0.2in
\end{figure*}

We document meta-validation accuracy plots for the CUB dataset in Figure \ref{fig:CUB_val} and the MiniImagenet dataset in Figure \ref{fig:MiniImagenet_val}. Respective plots but for the versions with $10$x higher inner loop learning rates can be found in Figures \ref{fig:CUB_biglr_val} and \ref{fig:MiniImagenet_biglr_val}. It can be quickly seen that in all plots except one (third row third column in Figure \ref{fig:MiniImagenet_biglr_val}) gradient sharing accelerates meta-training compared to the baseline. The acceleration effect is more pronounced in the $5$-task setting and less so in the $1$-task setting, which is not surprising because gradient sharing is a multi-task learning based inner loop regularizer. The $1$-task setting also occasionally results in a lower meta-validation accuracy peak compared to the baseline. This prompts important future work into inner loop regularizers that can strongly accelerate meta-learning while not sacrificing meta-test performance even in the absence of other tasks in the task batch. 

\subsection{Meta-Test Accuracy}
The legend in each of these meta-validation plots also indicates the maximum validation accuracy achieved, the meta-training epoch at which it was achieved, as well as the final meta-test accuracy. We note that the meta-test accuracies established for the MAML and MAML++ baselines generally reproduce or surpass what was reported in \citet{finn2017model}, \citet{antoniou2018train}, and \citet{antoniou2019learning}, even though specific hyperparameters might be slightly different. However, it seems that the baseline Meta-SGD meta-test accuracy often falls short of that of MAML, which is contrary to what was reported in \citet{li2017meta}. Like \citet{li2017meta}, we initialize all the entries of the vector learning rate $\boldsymbol{\alpha}$ to the same value. While we chose $0.1$ for fair comparison to MAML and MAML++, they mentioned that they randomly chose from $[0.005,0.1]$. It is possible that a hyperparameter search will enable Meta-SGD to outperform MAML, but we note that the meta-validation graphs indicate declining performance beyond a certain point, indicating the presence of task over-fitting. This happens more often than vanilla MAML, which makes sense because there are more inner loop parameters that can be over-fit, and thus, like our method, it would benefit from outer loop regularization.

\subsection{Momentum $m$ and Lambda $\lambda$ Variables}
\begin{figure*}[ht]
\vskip 0.2in
\begin{center}
\centerline{\includegraphics[width=\textwidth]{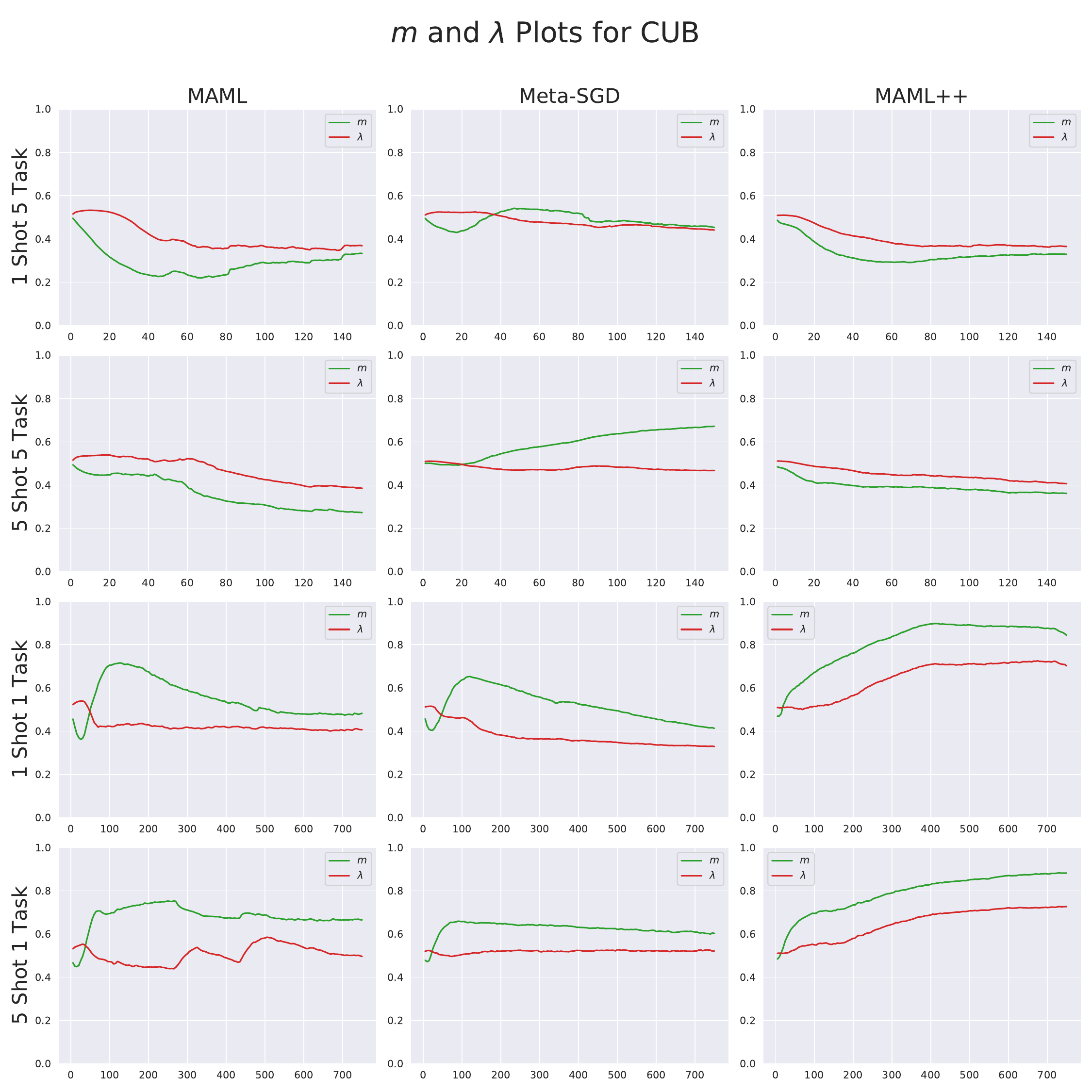}}
\caption{Evolution of Gradient Sharing Parameters throughout Meta-Training for the CUB dataset. The x axes denote the number of meta-training epochs, while the y axes denote the mean sigmoided value of the gradient sharing parameter. Specifically, $m$ denotes the average value of $\sigma(m_k)$ and $\lambda$ denotes the average value of $\sigma(\lambda_k)$ across $k \in [1,K]$. $K=5$ was set for all our experiments.}
\label{fig:CUB_momentum_avg}
\end{center}
\vskip -0.2in
\end{figure*}

\begin{figure*}[ht]
\vskip 0.2in
\begin{center}
\centerline{\includegraphics[width=\textwidth]{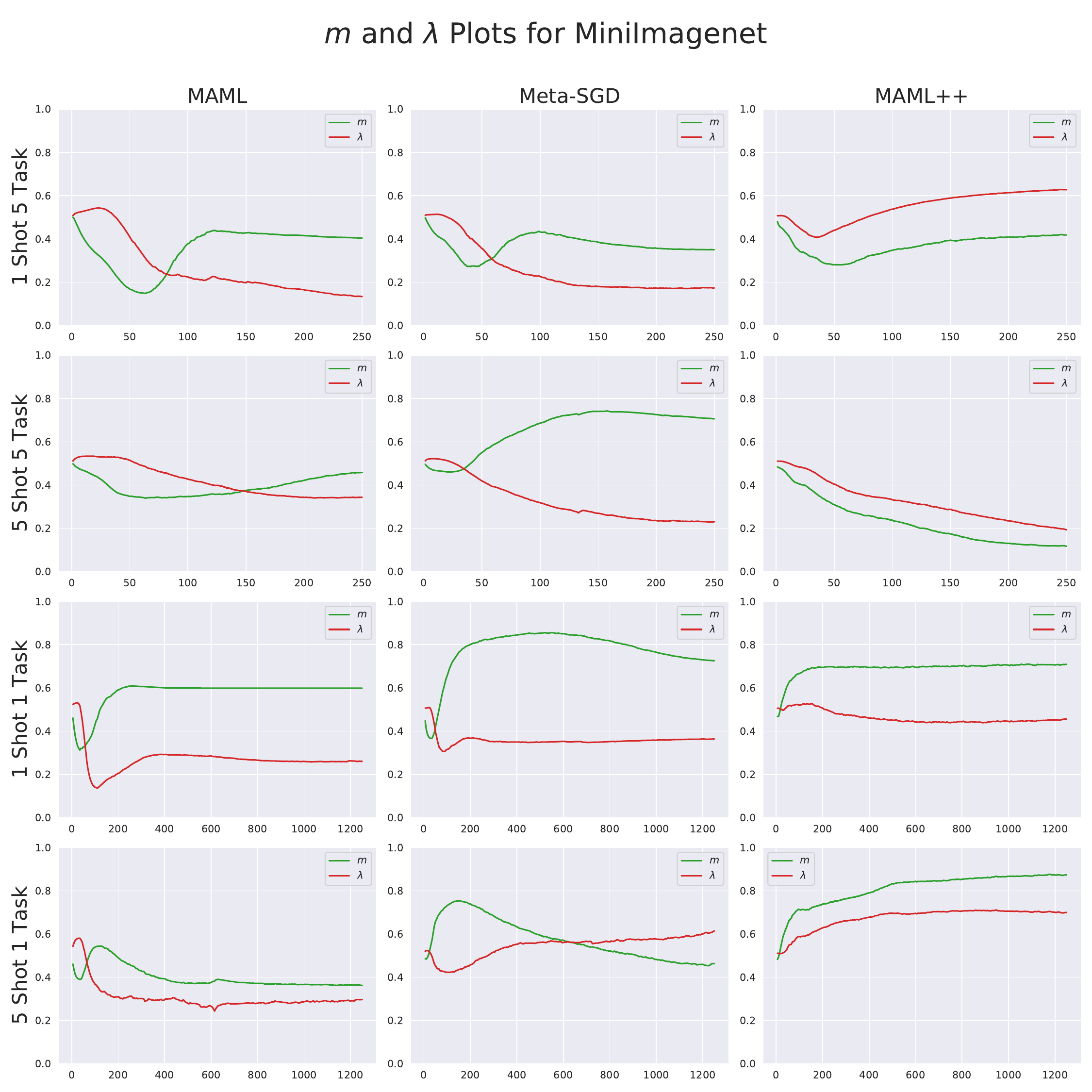}}
\caption{Evolution of Gradient Sharing Parameters throughout Meta-Training for the MiniImagenet dataset. The x axes denote the number of meta-training epochs, while the y axes denote the mean sigmoided value of the gradient sharing parameter. Specifically, $m$ denotes the average value of $\sigma(m_k)$ and $\lambda$ denotes the average value of $\sigma(\lambda_k)$ across $k \in [1,K]$. $K=5$ was set for all our experiments.}
\label{fig:MiniImagenet_momentum_avg}
\end{center}
\vskip -0.2in
\end{figure*}

\begin{figure*}[ht]
\vskip 0.2in
\begin{center}
\centerline{\includegraphics[width=\textwidth]{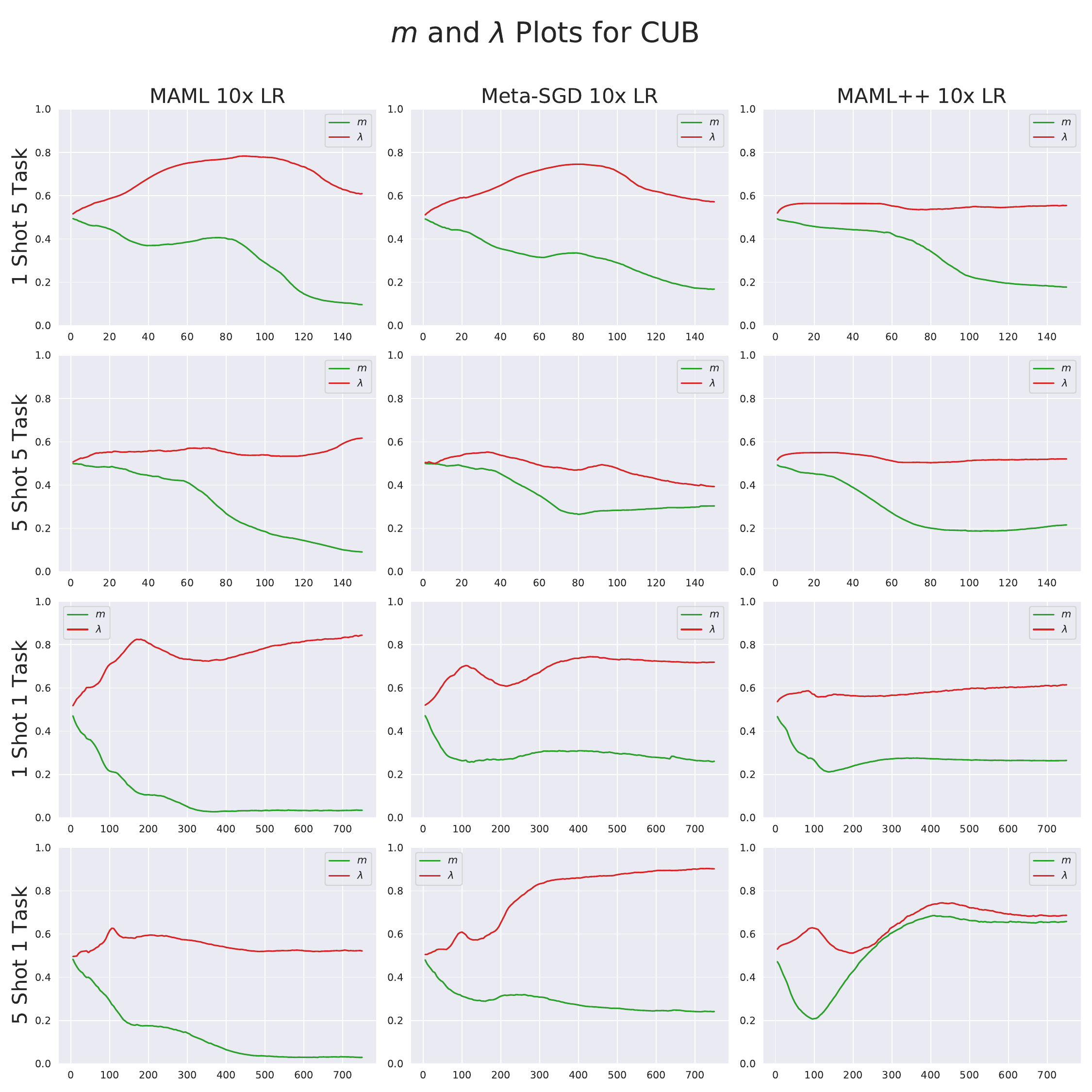}}
\caption{Evolution of Gradient Sharing Parameters throughout Meta-Training for the CUB dataset with $10$x the Inner Loop Learning Rate. The x axes denote the number of meta-training epochs, while the y axes denote the mean sigmoided value of the gradient sharing parameter. Specifically, $m$ denotes the average value of $\sigma(m_k)$ and $\lambda$ denotes the average value of $\sigma(\lambda_k)$ across $k \in [1,K]$. $K=5$ was set for all our experiments.}
\label{fig:CUB_biglr_momentum_avg}
\end{center}
\vskip -0.2in
\end{figure*}

\begin{figure*}[ht]
\vskip 0.2in
\begin{center}
\centerline{\includegraphics[width=\textwidth]{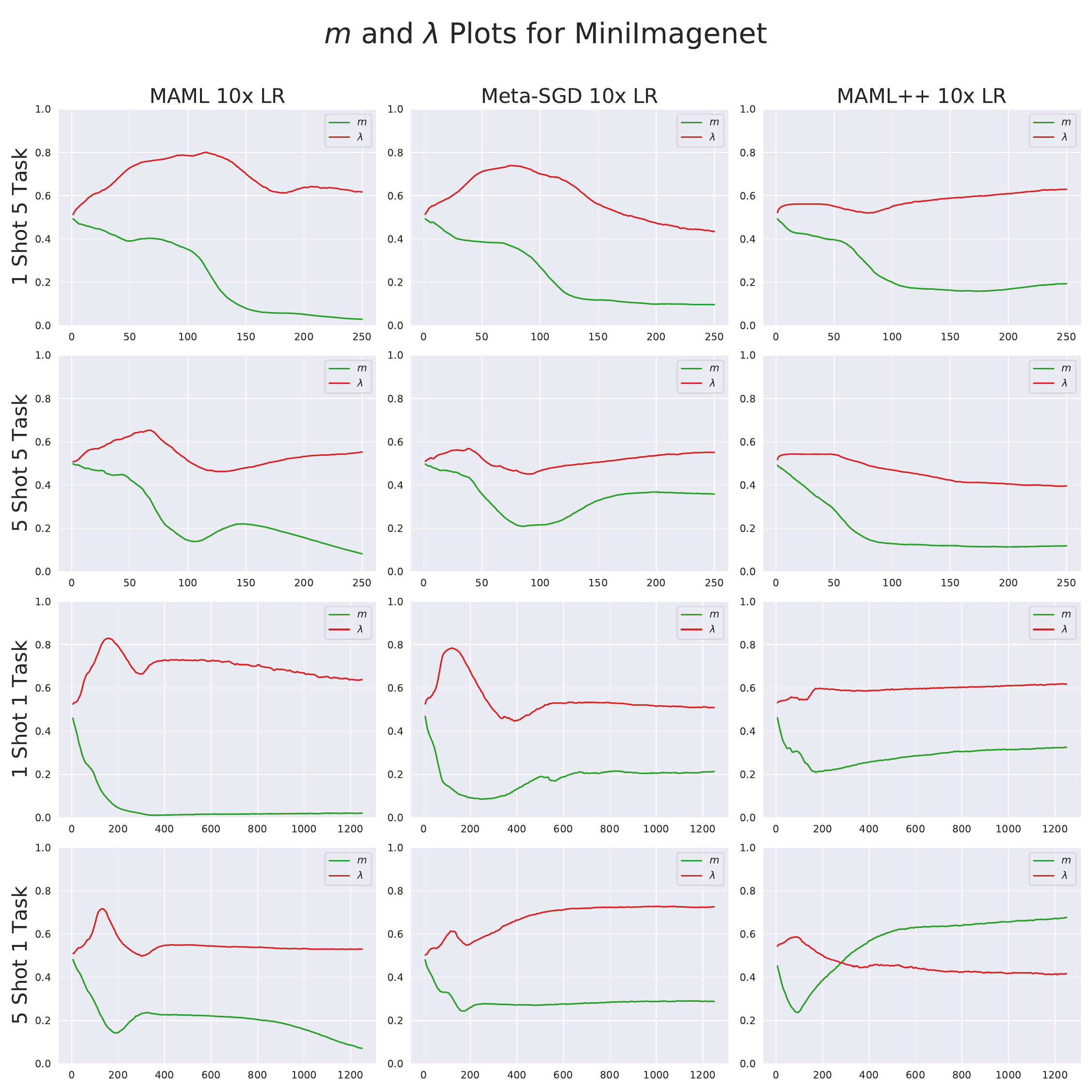}}
\caption{Evolution of Gradient Sharing Parameters throughout Meta-Training for the MiniImagenet dataset with $10$x the Inner Loop Learning Rate. The x axes denote the number of meta-training epochs, while the y axes denote the mean sigmoided value of the gradient sharing parameter. Specifically, $m$ denotes the average value of $\sigma(m_k)$ and $\lambda$ denotes the average value of $\sigma(\lambda_k)$ across $k \in [1,K]$. $K=5$ was set for all our experiments.}
\label{fig:MiniImagenet_biglr_momentum_avg}
\end{center}
\vskip -0.2in
\end{figure*}

Plots for $\boldsymbol{m}$ and $\boldsymbol{\lambda}$ are also documented in Figure \ref{fig:CUB_momentum_avg} for CUB and Figure \ref{fig:MiniImagenet_momentum_avg} for MiniImagenet, respectively Figures \ref{fig:CUB_biglr_momentum_avg} and \ref{fig:MiniImagenet_biglr_momentum_avg} for the versions with $10$x inner loop learning rate. If we compare them side-by-side with the meta-validation plots, it is easy to confirm our observation in Appendix Section \ref{section:more_analysis} that exemplary outcomes of gradient sharing correspond to low meta-learned $\boldsymbol{m}$ and $\boldsymbol{\lambda}$, while pathological outcomes correspond to high meta-learned $\boldsymbol{m}$ and $\boldsymbol{\lambda}$. In general, even for the pathological experiments, we can use the meta-validation set to perform early stopping and pick models that have yet to over-fit, so this is not a major issue.

\end{document}